\documentclass[3p, twocolumn, 11pt, times]{elsarticle}
\usepackage{flushend}
\usepackage{microtype}

\usepackage[l2tabu,orthodox]{nag}

\usepackage[colorlinks]{hyperref}

\usepackage[printonlyused]{acronym}

\usepackage{siunitx}
\usepackage[all]{nowidow}

\usepackage[dvipsnames]{xcolor}

\usepackage{epstopdf}

\usepackage{graphicx}
\usepackage{adjustbox}

\usepackage{import}
\usepackage{subcaption}

\graphicspath{{./latexGoodPractices/}}

\usepackage{booktabs}

\usepackage{tabu}

\usepackage{tabularx} 
\usepackage{multirow} 
\usepackage{array} 
\newcolumntype{Y}{>{\centering\arraybackslash}X} 

\usepackage{amssymb,amsfonts,amsmath,amscd}

\usepackage{bm} 

\newcommand{\bbm}{\begin{bmatrix}}
\newcommand{\ebm}{\end{bmatrix}}

\begin{document}

\begin{frontmatter}

\title{CropCraft: A Procedural World Generator for Robotic Simulation of\\Agricultural Tasks}

\author{Riccardo Bertoglio}
\author{Cyrille Pierre}
\author{Johann Laconte}
\author{Roland Lenain}

\affiliation{organization={Université Clermont Auvergne, INRAE, UR TSCF}, 
            city={63000, Clermont-Ferrand},
            country={France}}

\begin{abstract}
The adoption of agroecological practices in modern agriculture requires robotic systems capable of operating in highly diverse and complex field environments.
Developing and evaluating such systems relies heavily on simulation, yet generating realistic and configurable 3D environments representative of agroecological diversity remains a major challenge.
This paper presents CropCraft, an open-source procedural world generator built on Blender and Python, designed to produce 3D simulation environments tailored to agricultural robotics.
CropCraft generates crop fields from a simple YAML configuration file, supporting a wide range of scenarios including intercropping, vineyards, and weed-infested fields.
The tool includes a library of 3D plant models (crops, grasses, and weeds) at multiple growth stages, and uses stochastic placement algorithms to realistically reproduce the spatial variability observed in real fields.
Generated worlds are directly importable into the Gazebo simulator and include ground-truth annotations for all placed elements, supporting both perception and navigation algorithm development.
To demonstrate the practical utility of CropCraft, we apply it to the task of crop-weed semantic segmentation using deep learning.
A dataset of 10,000 synthetic images of maize fields with varying weed densities, growth stages, and lighting conditions was generated and used to train several segmentation architectures.
Models trained exclusively on synthetic data achieve a sim-to-real gap of approximately 10\% mean Intersection over Union (mIoU) on real field images, outperforming previous state-of-the-art synthetic generation approaches.
We further show that combining even a few real images with synthetic data improves generalization across domains, providing new insights into the effective use of synthetic data for agricultural perception tasks.
\end{abstract}





\end{frontmatter}


\section{Introduction}
	The reduction of environmental impact has become a major concern for our society, since pollution and chemical use associated with human activities are increasingly recognized not merely as contributors to climate change, but as sources of health risks for consumers, agricultural workers and ecosystems.
For example, the widespread use of pesticides and fertilizers can degrade soil and water quality, contaminate food chains and lead to chronic exposures that harm human health~\cite{zhou_comprehensive_2025}.
Conventional agricultural practices contribute significantly to these problems.
The heavy reliance on synthetic fertilizers and pesticides leads to soil degradation, biodiversity loss, and both environmental pollution and human health hazards via contaminated air, water, food and soil.
These negative impacts highlight the urgent need for substantial changes in food-production systems.
Although some shifts have already begun (for example with the adoption of Organic Agriculture) such approaches alone may not be sufficient to address the full complexity of environmental and health challenges.
A growing body of research supports the broader adoption of diversified, agroecological systems based on ecological principles rather than high chemical inputs~\cite{anderson_agroecology_2021}. 
These agroecological principles often involve combining different types of crops and species with practices like crop diversification, intercropping, use of cover crops and legumes, reduced or no-till, and organic amendments.
These practices can enhance soil health, increase biodiversity, reduce dependence on synthetic inputs, and improve ecosystem resilience~\cite{blaix_agroecological_2026}.
Implementing such systems requires more precise and context-specific actions, for example differential treatment of crops depending on species, soil condition, or pest pressure, which may increase the complexity of farm operations compared to conventional uniform practices.
However, long-term studies of diversified and organic systems show that, over decades, these systems can maintain stable yields while greatly improving soil quality, biodiversity, carbon sequestration and profitability~\cite{krause_organic_2024}.

\begin{figure*}[t]
    \centering
    \includegraphics[width=\linewidth]{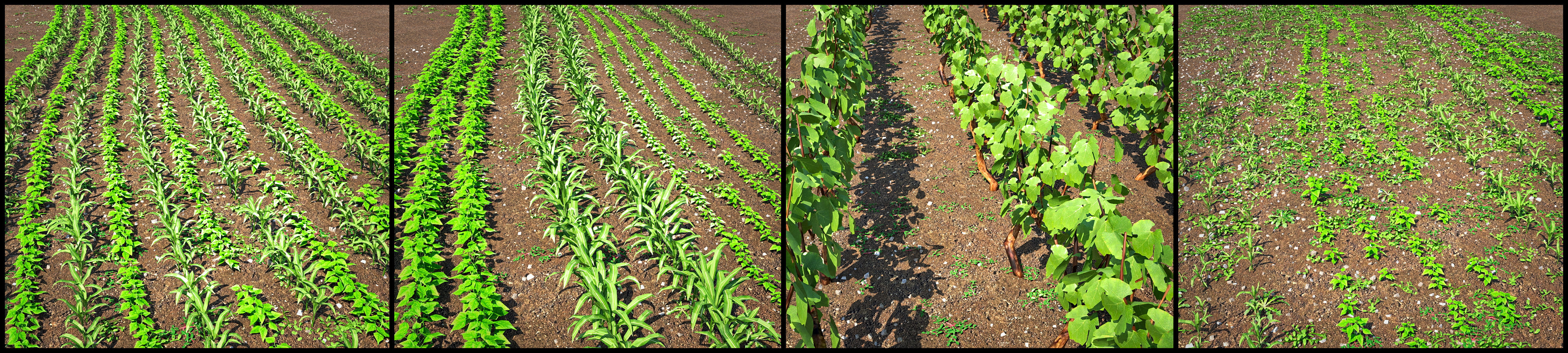}
    \caption{Examples of fields generated by the model (left to right): intercropped maize-bean system; well-defined maize and bean rows; a vineyard row; and a neglected field with missing crops and high weed density.}
    \label{fig:intro}
\end{figure*}

As a result, the large-scale application of agroecological principles requires the use of new tools capable of operating autonomously.
Agricultural robotics therefore emerges as a promising solution for promoting such agricultural practices~\cite{ditzler_automating_2022}.
To be effective, these robots must operate with a high level of accuracy when interacting with crops under diverse conditions.
The diversity of field configurations (e.g., market gardening, vineyards), types of crops, or weather conditions requires such robots to adapt their behaviors (perception and control algorithms), to the diversity of tasks and the varying contexts.
This imposes the development of adaptation mechanisms that need to be tested in various conditions, as well as environment recognition strategies to discriminate the nature of encountered vegetation.     

Research in environment recognition and scene interpretation in robotics has become very popular~\cite{crespo_semantic_2020}, as it enables robots to act on discriminated objects.
In agricultural fields, a key challenge lies in species discrimination, particularly for weeding applications~\cite{li_key_2022}, a topic that has attracted growing attention in agricultural robotics~\cite{avrin_design_2020}.
Often based on deep learning methods, such approaches face the challenge of requiring large amounts of annotated data (e.g., images and sensor acquisitions), which are difficult to generate and must be representative of the encountered conditions (sensor type, viewpoint, plant growth stage, weather conditions, etc.).
From a control perspective, a similar need arises, as navigation algorithms based on environment perception~\cite{pierre_multi-control_2022} must account for varying crop profiles.

The development of such robotic behaviors for agricultural applications can hardly rely solely on real-world experiments as it requires a large amount of annotated data.
As a result, studies in this area often make use of various simulation software tools~\cite{shamshiri_simulation_2018}.
Each of these tools has its own specificities in terms of representativeness, physics engines, visualization, and rendering.
Nevertheless, regardless of the simulation platform chosen, a major concern remains environment generation, which must be representative of both the application requirements and the diversity of possible situations.
In the context of robotics for agroecology, simulation environments must, for instance, handle mixtures of different crops, multiple types of weeds and grasses, the possible presence of non-vegetative elements (such as stones), as well as variations in ground type and soil properties.

Several types of environment generation have been proposed in the literature in the context of agriculture.
Most of them are dedicated to plant recognition and focus on image generation using AI~\cite{fawakherji_multi-spectral_2021}.
However, such tools do not enable the simulation of a robot navigating through the environment.
Three-dimensional world generation is addressed, for instance, in~\cite{shamshiri_simulation_2018} and~\cite{di_cicco_automatic_2017}.
The former proposes a generic generator that is not well suited to the specific requirements of agroecology, while the latter focuses on plant recognition from images and therefore appears limited for autonomous navigation or treatment applications relying on other sensors, such as 3D lidar or RGB-D cameras.
These examples nevertheless highlight the interest of 3D worlds that can be integrated into simulation software to support prototyping and development, as proposed in~\cite{robert_virtual_2020}, which is dedicated to simulation of the Oz platform.

In this paper, an environment generator named CropCraft\footnote{\url{https://github.com/Romea/cropcraft}} is proposed to create multiple types of environments representative of this diversity.
Its objective is to generate three-dimensional worlds representative of agroecological practices in the context of robotic development, including perception, recognition, and control.
CropCraft is a Python-based tool built on the open-source software Blender, enabling the generation of 3D environments that can be imported into the most common simulation platforms such as Gazebo, on which the world generation has been primarily tested.
CropCraft includes a library of plant models (crops, grasses, and weeds) at several growth stages, as well as various non-vegetative elements commonly found in fields, such as stones or ground irregularities.
The placement of these elements is performed automatically based on a user-defined configuration file that specifies parameters such as the number of crop rows, crop types, growth stages, geometric variability, probabilities of missing plants, levels and spatial distributions of weed infestation, and the presence, density, and variability of external elements.
Based on this configuration, a virtual environment (of which some examples are illustrated in Fig.~\ref{fig:intro}) is generated, capturing the variability and uncertainties encountered in real agricultural fields and directly importable into simulation software.
Since the positions of all elements are known, the generated worlds can be easily used for annotation purposes.
The tunable variability allows for evaluating the robustness of perception algorithms, while shape variations and missing-crop probabilities enable the analysis of control algorithm accuracy.
The environment generation relies on a library of lightweight models, ensuring compatibility with real-time simulation.

CropCraft can be used for a wide range of applications in agricultural robotics, including the development and evaluation of perception algorithms for crop and weed recognition, as well as the testing of navigation and control strategies under diverse field conditions.
Since its release, CropCraft has attracted interest from the research community and has been used in several projects.
For instance, \citet{liu_towards_2025} used CropCraft to build realistic simulated crop fields for testing their navigation algorithms.
The authors ran their lidar-based crop-row detection and robot navigation system within this simulated environment before validating it in real fields.

To demonstrate the versatility of CropCraft, this paper presents an example of its use in developing a crop-weed segmentation algorithm based on deep learning.
We leveraged CropCraft to generate synthetic images (Figure \ref{fig:synthetic_examples}) from 3D models of crop rows containing various types of weeds.
These images were then used to train deep learning models to distinguish crops from weeds, and their accuracy was compared with models trained on real images.
We show that our generation pipeline achieves a sim-to-real gap of 10\%, which is lower than previous state-of-the-art approaches \cite{di_cicco_automatic_2017}, which reported a sim-to-real gap of approximately 20\% (mIoU).
Furthermore, we provide new insights into effectively combining synthetic and real images for model training.
 
In summary, the main contributions are:
\begin{enumerate}
    \item The presentation of CropCraft, an open-source software tool for generating 3D environments representative of agroecological practices, designed to support the development of robotic applications in agriculture, made publicly available at \url{https://github.com/Romea/cropcraft}
    \item An application of CropCraft for generating synthetic images for crop-weed segmentation, demonstrating its effectiveness in reducing the sim-to-real gap and providing insights into training deep learning models with synthetic and real images. The associated training code is made publicly available at \url{https://github.com/Romea/crop-weed-segmentation}.
    \item A synthetic dataset of 10,000 images of maize plants and weeds, generated using CropCraft, showcasing a wide variety of field conditions including different weed densities, growth stages, and lighting conditions, made publicly available at \url{https://doi.org/10.57745/TNCSLP}.
\end{enumerate}

The paper is organized as follows.
Section 2 reviews related works, highlighting the contributions of CropCraft in comparison with existing simulation tools and environment generation methods in the context of agricultural robotics.
Section 3 is dedicated to a detailed description of the open-source software CropCraft, including its architecture, features, and configuration options.
Section 4 presents an example application of the software for the development of a robotic task, illustrating its use in generating synthetic images for training crop-weed identification algorithms.
Finally, Section 5 concludes the paper and discusses future directions, including potential extensions to model vegetation deformation, which are not addressed in the current version of CropCraft.

\section{Related Work}
	To successfully deploy robots in real-world environments, it is crucial that robotic systems undergo extensive testing across a wide range of conditions to ensure their resilience and safety. 
Although theoretical proofs offer critical insights into system behavior, they inherently depend on assumptions regarding noise distribution, robotic behavior, and environmental factors (e.g.,~\cite{laconteCertifyingMapsSafe2024}).
Consequently, empirical testing through repeated experiments under real-world conditions remains an indispensable step to validate the reliability of these systems.
This need for comprehensive testing is particularly crucial in the context of agricultural robotics, where robots are poised to replace the most tedious and laborious tasks in the field. 
For these robots to be truly effective, they must operate year-round in dynamic and constantly changing environments. 
Furthermore, they must be versatile enough to handle a variety of tasks, from weeding to managing crops of different types and sizes.

To support the development of such systems, various datasets have been made available to the research community.
In the context of SLAM and localization, \citet{soncini_rosario_2025} proposed the Rosario dataset, consisting of agricultural scenes of soybean crops, highlighting challenges such as highly repetitive structures and exposure variations for vision-based algorithms.
Complementary resources include the BLT (BACCHUS Long-Term) dataset~\cite{polvara_bacchus_2024}, which emphasizes long-term autonomy through repeated traversals over extended periods and changing environmental conditions.
The BotanicGarden dataset~\cite{liu_botanicgarden_2024} focuses on structured yet biologically diverse garden environments, capturing seasonal appearance changes and complex vegetation geometry relevant for robustness evaluation in natural settings.

As such, while accurate localization is the cornerstone of every mobile robot application, an agricultural robot must also be capable of performing its assigned tasks, such as monitoring crops.
For example, phenotyping (i.e., the assessment of plant properties) is a labor-intensive task that robots could undertake.
To support this, \citet{schunckPheno4DSpatiotemporalDataset2021} provided a detailed dataset of tomato and maize point clouds, offering labeled data for plant segmentation and 3D reconstruction.
This dataset was collected in an indoor setting, resulting in clean and highly detailed point clouds.
Additionally, \citet{marks_bonnbeetclouds3d_2024} presented a dataset of sugar beet point clouds captured in the field using drones.
This dataset provides highly detailed representations of plants in real agricultural conditions, with each leaf labeled, offering an extensive resource for phenotyping in natural outdoor environments.

Weeding is another crucial task in agricultural robotics, requiring accurate detection and differentiation between crops and weeds to enable precise and effective intervention.
To address this challenge, several datasets have been developed for training and evaluating perception algorithms.
For example, the Sugar Beets 2016 dataset~\cite{chebrolu_agricultural_2017} provides annotated field images of sugar beet crops and associated weeds, supporting research in crop-weed segmentation under real agricultural conditions.
The Crop and Weed Dataset introduced by \citet{steininger_cropandweed_2023} is a large-scale image dataset specializing in the fine-grained identification of 74 relevant crop and weed species with a strong emphasis on data variability.
Similarly, The ACRE Crop-Weed Dataset provides high-resolution RGB images of maize and bean crops with four weed species, collected using a ground robot and annotated for instance segmentation \cite{bertoglio_acre_2025}.
Additionally, \citet{olsenDeepWeedsMulticlassWeed2019} provided a comprehensive multiclass image dataset specifically designed for weed classification models.

However, creating effective datasets for robotic applications requires extensive manual labeling and the capture of diverse data to fully represent the complexity of real-world environments.
In agriculture, this complexity includes variations in crop types, different growth stages, and a wide range of field and weather conditions.
While real-world datasets are invaluable for evaluating robotic algorithms, they inevitably leave some scenarios unexplored due to these limitations.
Given that deployed robots must meet stringent safety and performance standards, simulated data can play a crucial role in bridging these gaps.
For example, \citet{paigwarProbabilisticCollisionRisk2020} demonstrated the use of the CARLA simulator to validate robotic algorithms in the context of autonomous driving.
In this way, simulations allow for extensive testing that would be difficult or impossible to replicate in real-world settings.

In agricultural settings, simulations of the Oz agribot have demonstrated the effectiveness of virtual environments in identifying software issues that would otherwise be costly to detect through field testing~\cite{robert_virtual_2020}.
Similarly, the AgROS emulation tool provides a framework for evaluating the performance of autonomous ground vehicles in precision farming, enabling farmers to assess robotic operations in a quasi-real-world setting~\cite{tsolakisAgROSRobotOperating2019}.
Building on this, the concept of digital twins has been introduced in agricultural robotics through a multi-agent architecture designed for dynamic fleet simulations~\cite{gutierrezcejudoAgrirobotDigitalTwins2023}.
This approach enables real-time coordination and task allocation among multiple robots, thereby enhancing the scalability and realism of agricultural simulations.
Virtual reality simulators, such as those developed for precision agriculture machinery, further illustrate the potential of simulation to optimize the design and operation of agricultural equipment in an interactive and immersive manner~\cite{cutiniCoSimulationVirtualReality2023}.

In parallel, simulation and data synthesis have played a central role in advancing perception and vision systems in agricultural robotics by enabling the generation of large and diverse training datasets.
Procedural generation using 3D simulation platforms and game engines allows fine control over environmental factors such as lighting, weather, plant geometry, and growth stages, and has been successfully applied to produce synthetic crop-weed imagery for training segmentation and detection models~\cite{di_cicco_automatic_2017, carbone_augmentation_2022}.
Complementary approaches based on cut-and-paste techniques leverage real plant instances composited onto varied backgrounds, achieving competitive performance while reducing data collection efforts, albeit with limited control over environmental realism and a continued dependence on real annotated data~\cite{sapkota_use_2022, picon_deep_2022}.
More recently, generative AI methods, including GANs and diffusion models, have demonstrated strong capabilities for synthetic data augmentation, enabling the creation of diverse and realistic crop-weed images and, in some cases, reducing the need for manual labeling~\cite{chong_unsupervised_2023, chen_synthetic_2024}. 
Despite these advances, most existing approaches focus on camera-based perception and remain constrained by realism, controllability, computational cost, or reliance on real data.
Moreover, limited attention has been given to synthetic data generation tailored to range-based sensors.
To address this gap, we propose a novel synthetic world generator specifically optimized for real-time lidar simulation. 
This generator allows for on-the-fly configuration of virtual environments, enabling extensive and statistically robust evaluations of robotic algorithms in a variety of scenarios.

\section{Description}
	In this section, we present a detailed description of CropCraft.
First, we introduce the overall structure of the framework and the architecture of the generated fields.
Then, we explain how crops and weeds are generated using stochastic placement algorithms.
Finally, we provide a thorough description of the ground-truth information produced by the framework.

\subsection{Overall structure}

The aim of this generator is to create a 3D model of a crop field based on a configuration file that describes the different characteristics of the field.
To achieve this, the program relies on procedural generation tools developed in Blender, along with Python scripts for asset retrieval and configuration file parsing.
Using a YAML configuration file allows users to define agricultural fields in a simple text format without requiring any prior knowledge of 3D modeling software.
The type of output data is also configurable and depends on the intended application.
Currently, the framework supports the following options:
\begin{itemize}
\item Saving the Blender file, which allows users to open it in Blender for manual modifications after field generation and to export it manually.
\item Exporting the environment to the Gazebo Classic simulator, enabling integration into an existing simulation setup.
\item Creating a description file containing configuration data and detailed information about each generated crop.
\item Generating images and annotations from predefined camera poses, which can be used to build datasets for vision-based algorithms.
\end{itemize}

The configuration file defines the structure and characteristics of the generated field.
The field is composed of several beds, where the term \emph{bed} is used in accordance with its definition in the market gardening domain.
In the context of this article, it refers to the strip of land on which the agricultural vehicle operates.
Each bed can contain one or multiple rows of crops. Moreover, each bed can be configured independently, providing users with the flexibility to design various types of fields, including vineyards or conventional crop fields.

In addition to defining the crop layout, the configuration file includes parameters for weed and stone scattering.
This feature enables the evaluation of crop detection algorithms in a simulated environment.
Users can control the density and types of plants, allowing them to adjust the difficulty of perception tasks.
Furthermore, the option to add stones or soil clods enhances the level of detail of the 3D model, ensuring that the generated fields realistically reflect the complexity of real farming environments.

\subsection{Crops generation}

During the crop generation process, the framework allows users to define several parameters that govern the placement and characteristics of crops within each bed.
Users can specify the number of rows, the distance between rows, the spacing between individual plants within a row, and the overall width of the bed.
This level of customization is essential for accurately modeling various agricultural practices and ensuring that the generated fields reflect real-world scenarios.

To enhance realism, the program applies Gaussian noise to the computed poses of the plants.
Users can define the standard deviation of this noise for both position and orientation.
Orientation noise is applied to the roll and pitch axes, while the yaw angle is sampled from a uniform distribution in the range \SIrange{0}{360}{\degree}.
For vineyard applications, this random yaw selection can be disabled, as grapevines typically grow with a fixed orientation aligned with trellising systems.

Additionally, the framework supports the generation of curved trajectories, accommodating agricultural plots that naturally follow the contours of the field edges.
Users can specify a mathematical function in the configuration file to define this curvature, allowing for a more accurate representation of real farming layouts.
The function is expressed as $y = f(x)$, where $x$ represents the position along the length of the bed and $y$ denotes the lateral offset.

\begin{figure*}[t]
    \centering
    \begin{subfigure}[t]{0.49\textwidth}
        \centering
        \includegraphics[width=\linewidth]{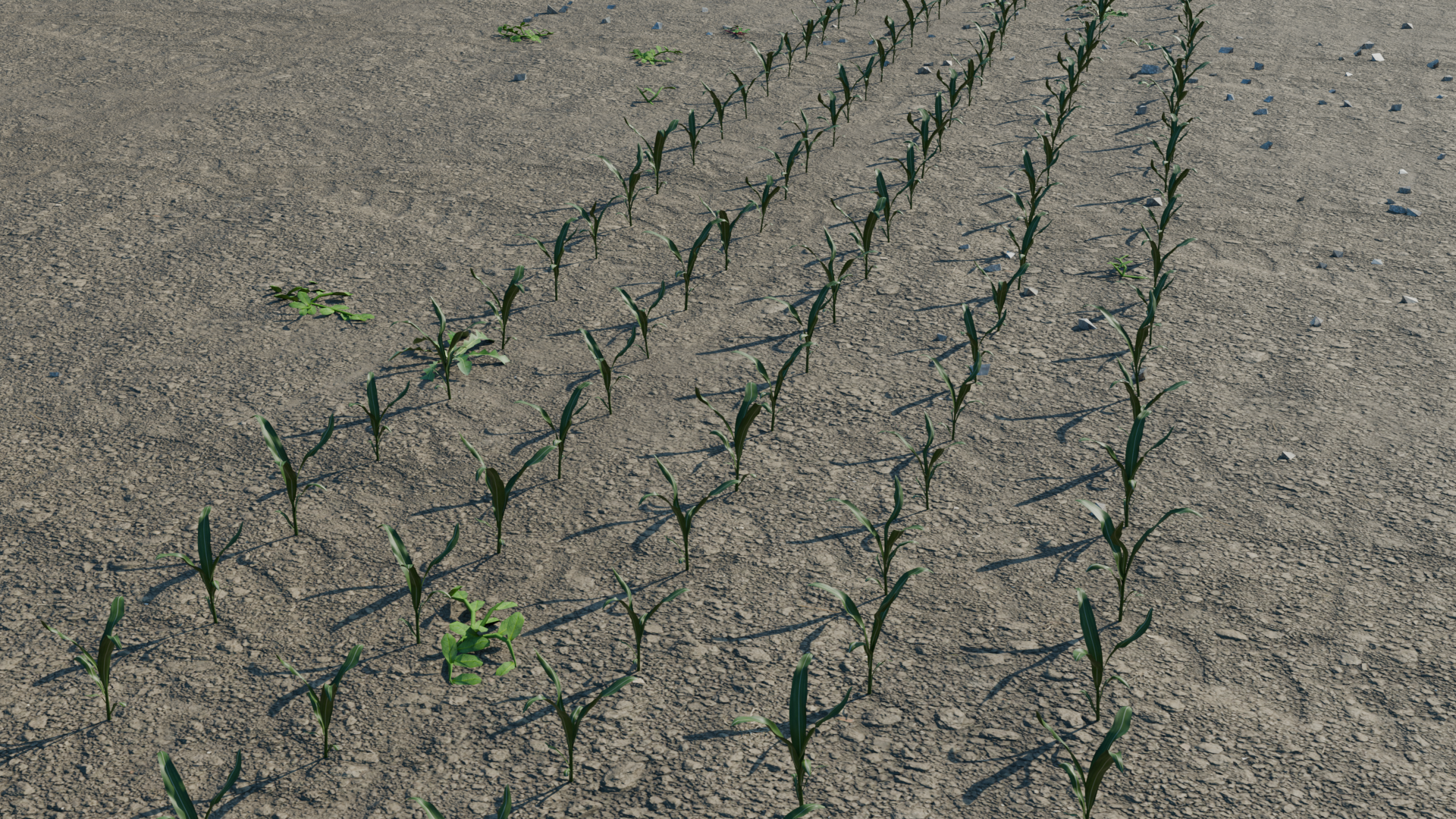}
        \caption{Straight rows of small maize plants with scattered weeds.}
    \end{subfigure}%
    \hfill
    \begin{subfigure}[t]{0.49\textwidth}
        \centering
        \includegraphics[width=\linewidth]{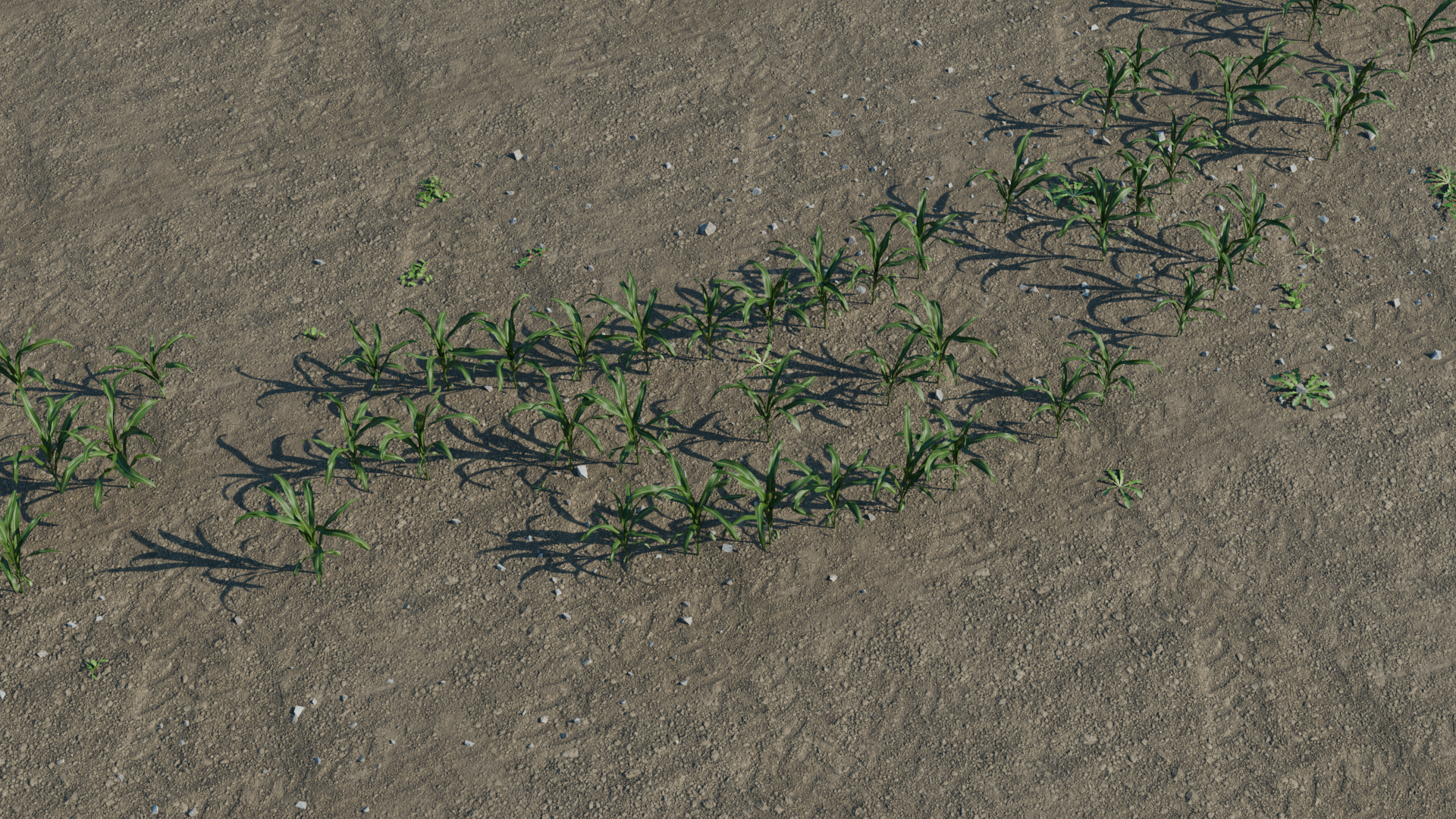}
        \caption{Later growth stage with curved bed layout and missing plants.}
    \end{subfigure}
    \caption{Example outputs from CropCraft crop generation. The left image shows early-stage straight maize rows with weeds, while the right image demonstrates curved row generation, taller plants, and occasional missing plants.}
    \label{fig:crops_generation}
\end{figure*}

Finally, users should provide a set of existing 3D models representing the desired plants to be used in the beds.
The program allows users to specify an approximate height for the plants in each bed and then selects a subset of models that closely match this height from the available assets.
Once selected, the models are resized to conform to the specified height, with log-normal noise applied to the scaling factor to introduce natural variation in plant sizes.
This approach occasionally produces plants that deviate significantly from the target height, reflecting the diversity observed in real crop fields.
Figure~\ref{fig:crops_generation} illustrates two generated maize setups: one with straight rows and early-stage plants, and another with curved planting lines, taller plants, and occasional missing plants.

A dedicated parameter also controls the probability of a plant position being left empty, simulating the missing-plant phenomenon commonly encountered in real fields due to germination failures or mechanical damage.

\subsection{Weed generation}

The aim of the weed generation module is to produce a realistic spatial distribution of weeds across the field.
A naive approach based on a uniform random distribution would not be representative of real conditions: as observed by \citet{hamouzFieldscale2004}, weed populations in agricultural fields tend to be spatially heterogeneous, forming clusters of varying density rather than being uniformly spread.

To reproduce this behavior, CropCraft uses a density-map-based placement strategy combined with Poisson disk sampling, implemented using Blender's Geometry Nodes.
The density map is generated procedurally using a heterogeneous terrain noise function (a variant of fractional Brownian motion), which produces spatially correlated patterns with natural-looking spatial variation.
The noise function operates in 4D space, where the fourth dimension encodes a seed value; this allows the generated map to be varied reproducibly by changing the seed while keeping all other parameters fixed.
Each weed species is scattered independently, with its own density map, allowing different species to occupy distinct spatial niches within the same field.

Plant instances are then distributed over the ground surface using Poisson disk sampling, which enforces a minimum distance between neighboring plants.
This avoids the unnatural clustering produced by purely random placement while still respecting the density modulation from the noise map.
For each sampled position, a 3D plant model is selected at random from the pool of available assets for the corresponding weed species.
Each instance is assigned a random yaw rotation drawn uniformly from \SIrange{0}{360}{\degree}, along with a random scale factor to introduce natural variation in plant size.

As an alternative to procedural noise, CropCraft also supports image-based density maps: the user can provide a custom image file whose pixel intensities directly define the local weed density.
This mode replaces the noise generation step and allows the simulation of specific spatial infestation patterns, for instance derived from field surveys or agronomic models.

\begin{figure*}[t]
    \centering
    \begin{subfigure}[t]{0.31\textwidth}
        \centering
        \includegraphics[height=0.5625\linewidth]{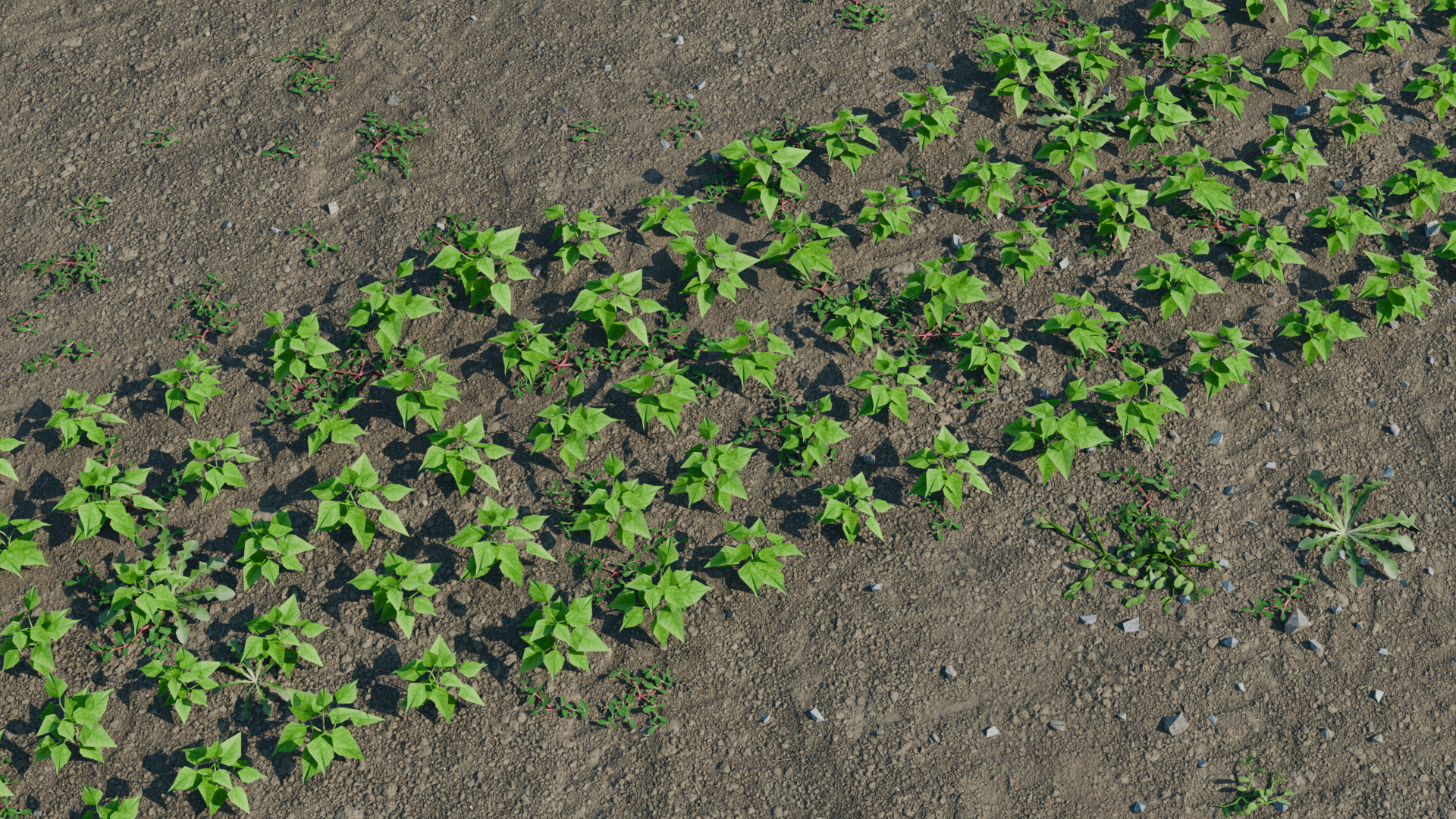}
        \caption{Bean rows with dense portulaca weeds and lower-density polygonum and taraxacum.}
    \end{subfigure}%
    \hfill
    \begin{subfigure}[t]{0.31\textwidth}
        \centering
        \includegraphics[height=0.5625\linewidth]{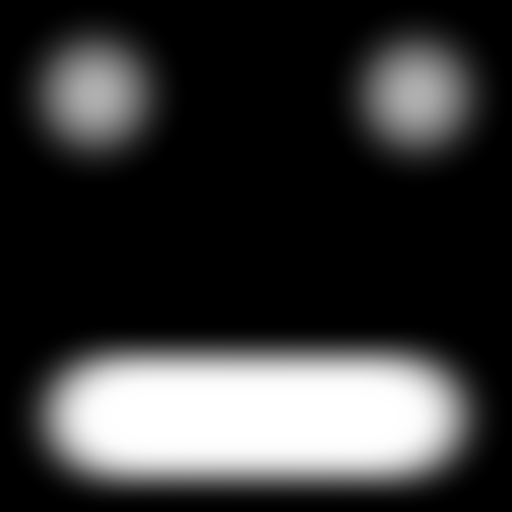}
        \caption{Image-based density map used to control weed placement.}
    \end{subfigure}%
    \hfill
    \begin{subfigure}[t]{0.31\textwidth}
        \centering
        \includegraphics[height=0.5625\linewidth]{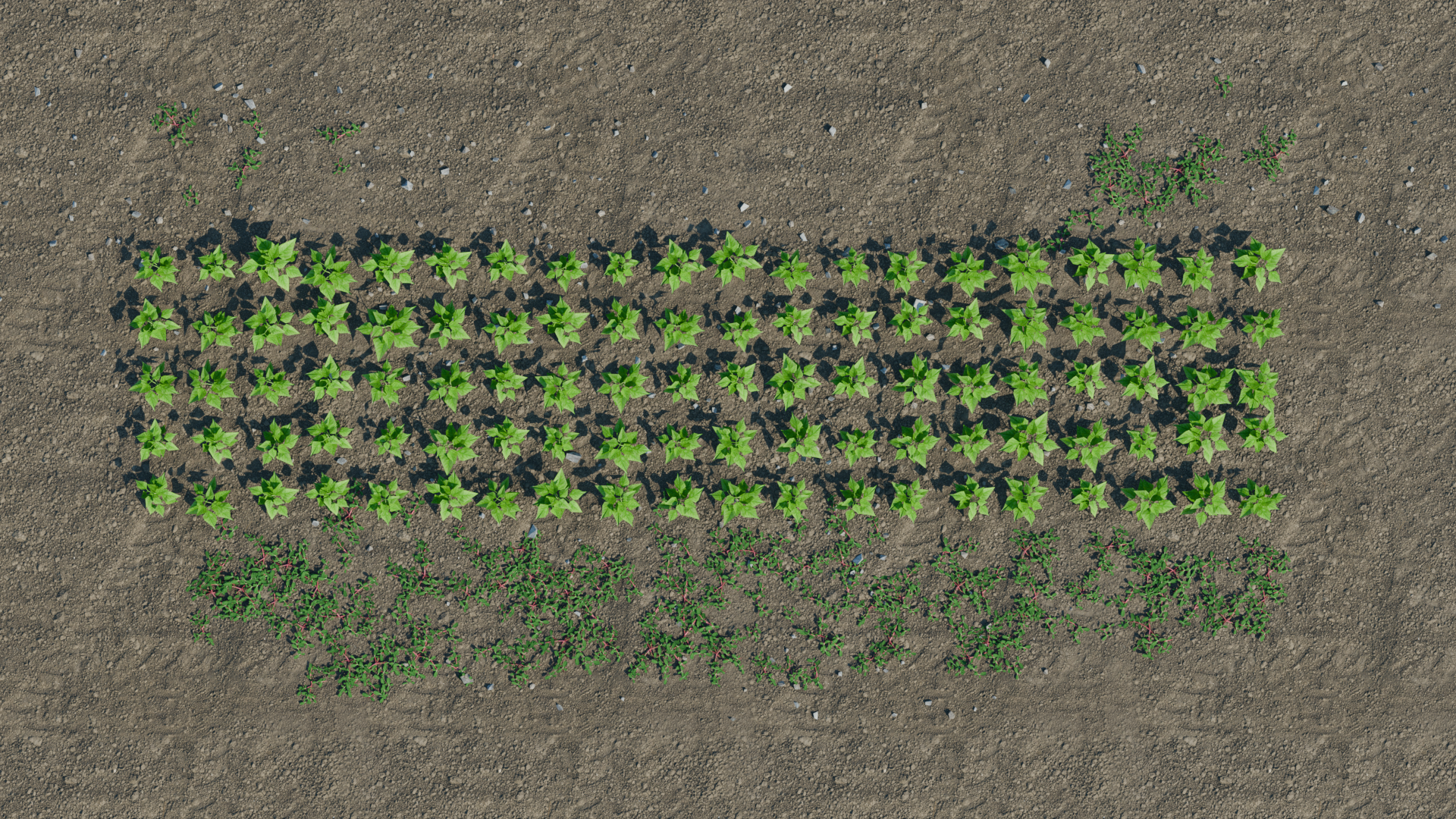}
        \caption{Bean rows with weeds generated from the provided noise pattern.}
    \end{subfigure}
    \caption{Weed generation examples from CropCraft. The left image shows stochastic weed placement with mixed species densities; the center image is the external density map; the right image shows weeds placed according to that pattern.}
    \label{fig:weed_generation}
\end{figure*}

Figure~\ref{fig:weed_generation} shows two bean field examples generated by CropCraft.
The left image uses only the internal stochastic placement strategy to produce high-density portulaca patches interspersed with lower-density polygonum and taraxacum clusters.
The right image uses the external density pattern shown in the center panel, demonstrating how an arbitrary image can drive weed distribution.

The configuration exposes the following parameters per weed species: the plant type, the maximum plant height, the desired density (in plants per unit area), the minimum distance between instances (Poisson disk radius), and the noise scale and offset controlling the shape and intensity of the density map.
Additionally, non-vegetative elements such as stones or soil clods can be scattered using the same mechanism, further increasing the realism of the generated environment.

\subsection{Ground-truth production}

A key feature of CropCraft is its ability to produce ground-truth data alongside the generated 3D environment.
Since all plant poses and model identities are determined during generation, the framework has complete knowledge of the scene and can export this information in several complementary forms.

\paragraph{Field description file}
CropCraft exports a structured description of the generated field containing, for each crop plant: its 3D position and orientation (roll, pitch, yaw), the 3D model used, and the associated metadata (height, width, and leaf area).
The data are organized hierarchically by bed and row, and are available in JSON format for human readability or in MessagePack binary format (optionally gzip-compressed) for compact storage and fast loading.
This file enables validation of algorithms that reason about field structure, such as detecting missing plants in a row or estimating plant density. It also supports the computation of agronomic metrics, such as the total leaf area of a row, a bed, or the entire field.

\paragraph{Lidar ground truth}
When exporting to Gazebo, each object in the scene is annotated with a \texttt{laser\_retro} value embedded in its SDF collision description.
This value is read by Gazebo's lidar sensor plugin and modulates the intensity channel of the returned point cloud.
The encoding is as follows: the ground surface is assigned an intensity of $0$, each crop bed receives a positive integer equal to its index (i.e.,\ $1, 2, 3, \ldots$), each weed species is assigned a negative value equal to $-(i+1)$ where $i$ is its index, and non-vegetative elements such as stones are assigned $-0.5$.
This makes it possible to directly identify the type and origin of each Lidar return from the intensity channel, without any post-processing, and to benchmark point-cloud-based plant detection or segmentation algorithms against a point-accurate reference.

\begin{figure}[h]
    \centering
    \begin{subfigure}[b]{0.45\textwidth}
        \centering
        \includegraphics[width=\linewidth]{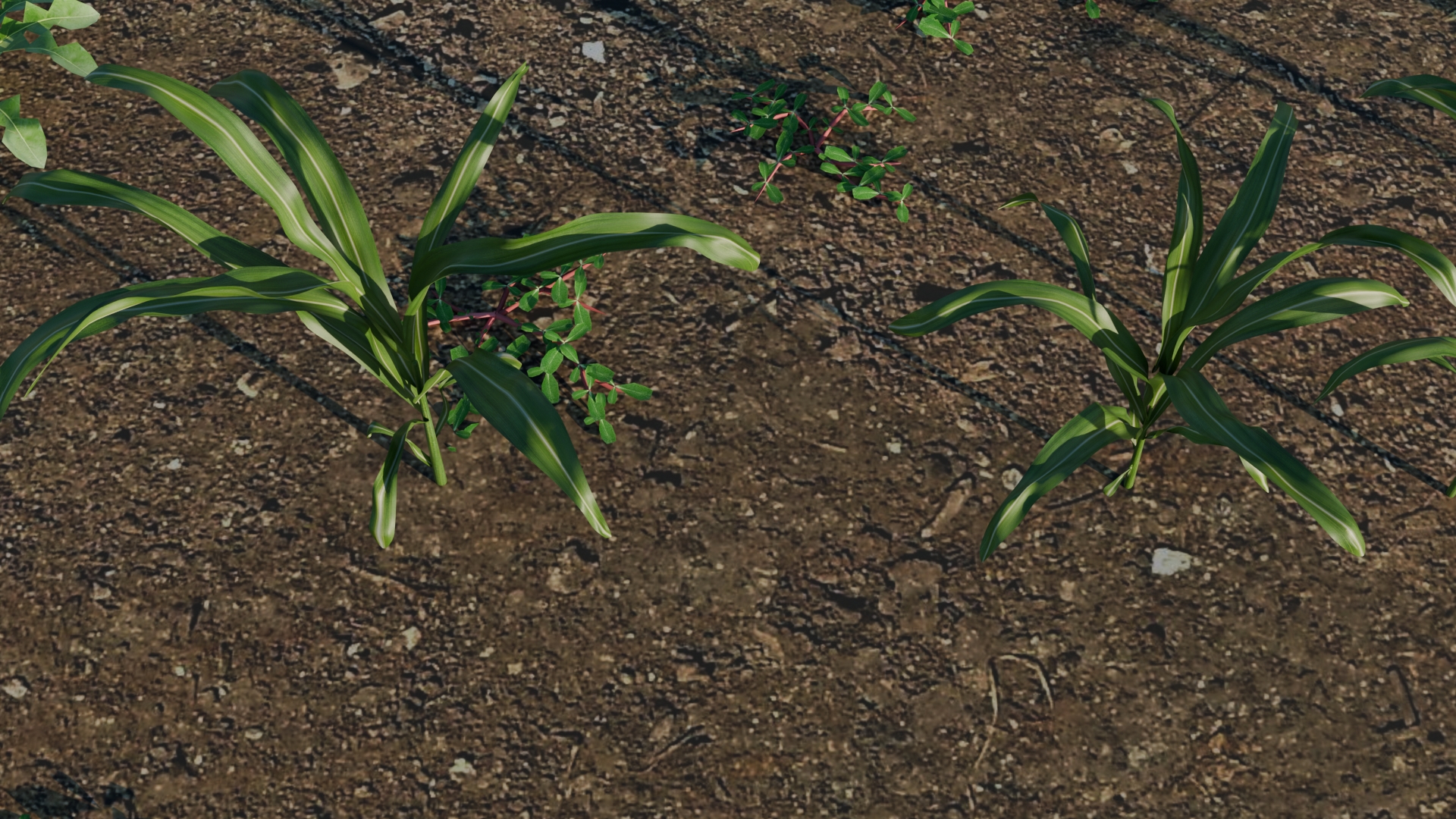}
        \caption{RGB image of maize plants with weeds.}
    \end{subfigure}%
    \hfill
    \begin{subfigure}[b]{0.45\textwidth}
        \centering
        \includegraphics[width=\linewidth]{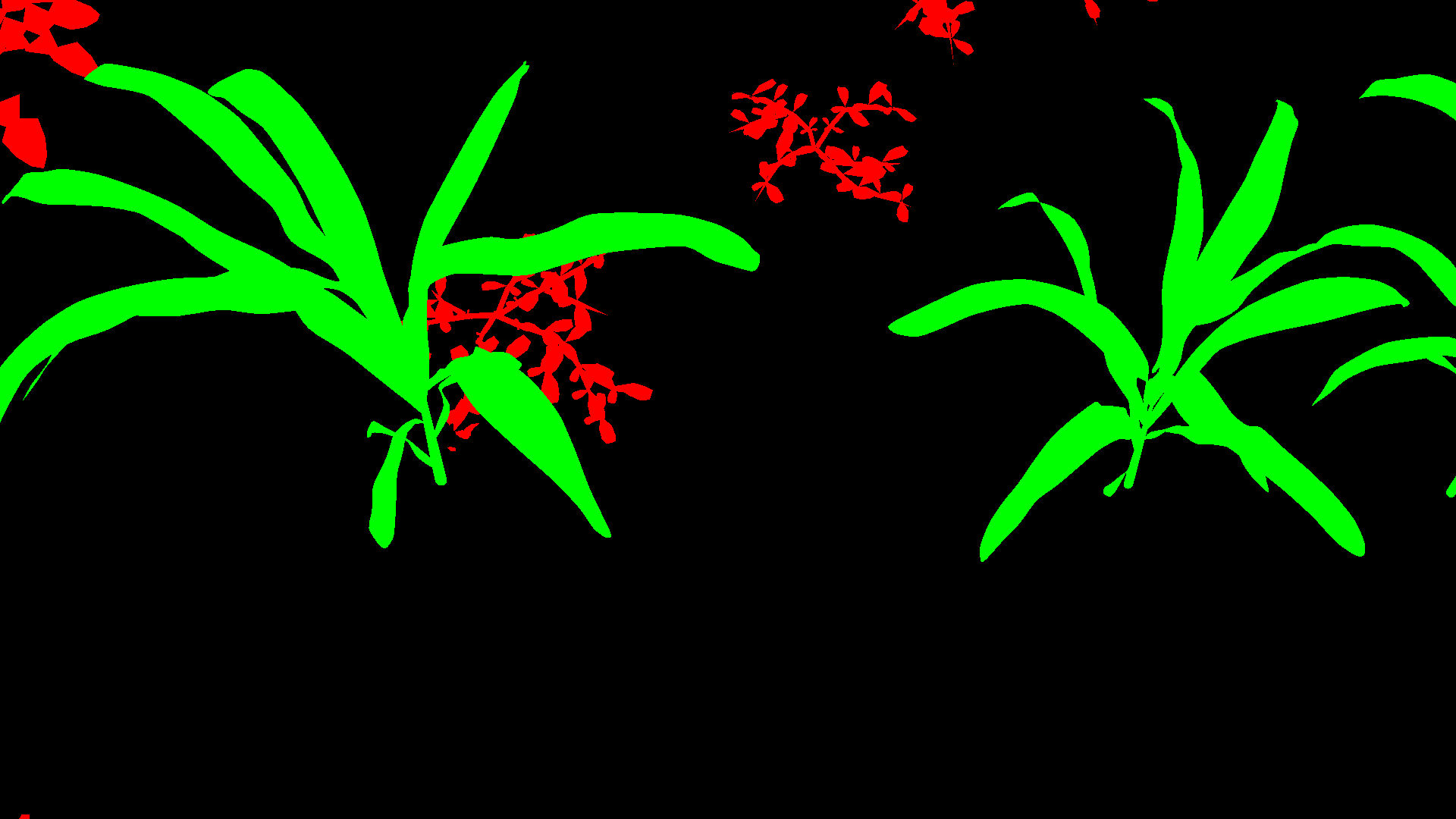}
        \caption{Semantic segmentation mask: crop (green), weeds (red), background (black).}
    \end{subfigure}
    \caption{Example of an image-mask pair generated by CropCraft, showing maize plants with weeds and the corresponding semantic segmentation labels.}
    \label{fig:segmentation_example}
\end{figure}

\paragraph{Camera segmentation masks}
CropCraft can render semantic segmentation masks for any set of user-defined camera poses.
The masks are generated by rendering the scene using emissive materials, where each semantic class (crop, weed, background) is assigned a distinct RGB color.
A nearest-color quantization step is applied after rendering to eliminate antialiasing artifacts and ensure that every pixel is assigned to exactly one class.
The masks are exported as PNG images and can be used directly to train or evaluate vision-based crop and weed recognition algorithms, as demonstrated in Section~\ref{sec:datasets}.
An example of a generated image-mask pair is shown in Figure~\ref{fig:segmentation_example}.

\section{Application}
	
As an example of application, we show how the tool can be used to generate synthetic images for training deep learning models for crop and weed semantic segmentation. 
We compare different deep learning architectures trained on synthetic images and evaluate their performance on real images collected in maize fields.
In Section \ref{sec:datasets} we present the synthetic and real-world datasets, detailing the image generation process and dataset characteristics.
In Section \ref{sec:exp_setup} we then outline the experimental design, including the training strategies and evaluation metrics adopted to assess model performance.
Finally, in Section \ref{sec:dl_training} we provide the training protocol and architecture details of the deep learning models tested, along with the data augmentation and preprocessing steps applied.

\subsection{Datasets}\label{sec:datasets}
\subsubsection{Synthetic Data}
\begin{figure}[t]
    \centering
    \subcaptionbox{Afternoon, tall maize, low weed density\label{fig:a}}[0.48\columnwidth]{
        \includegraphics[width=\linewidth]{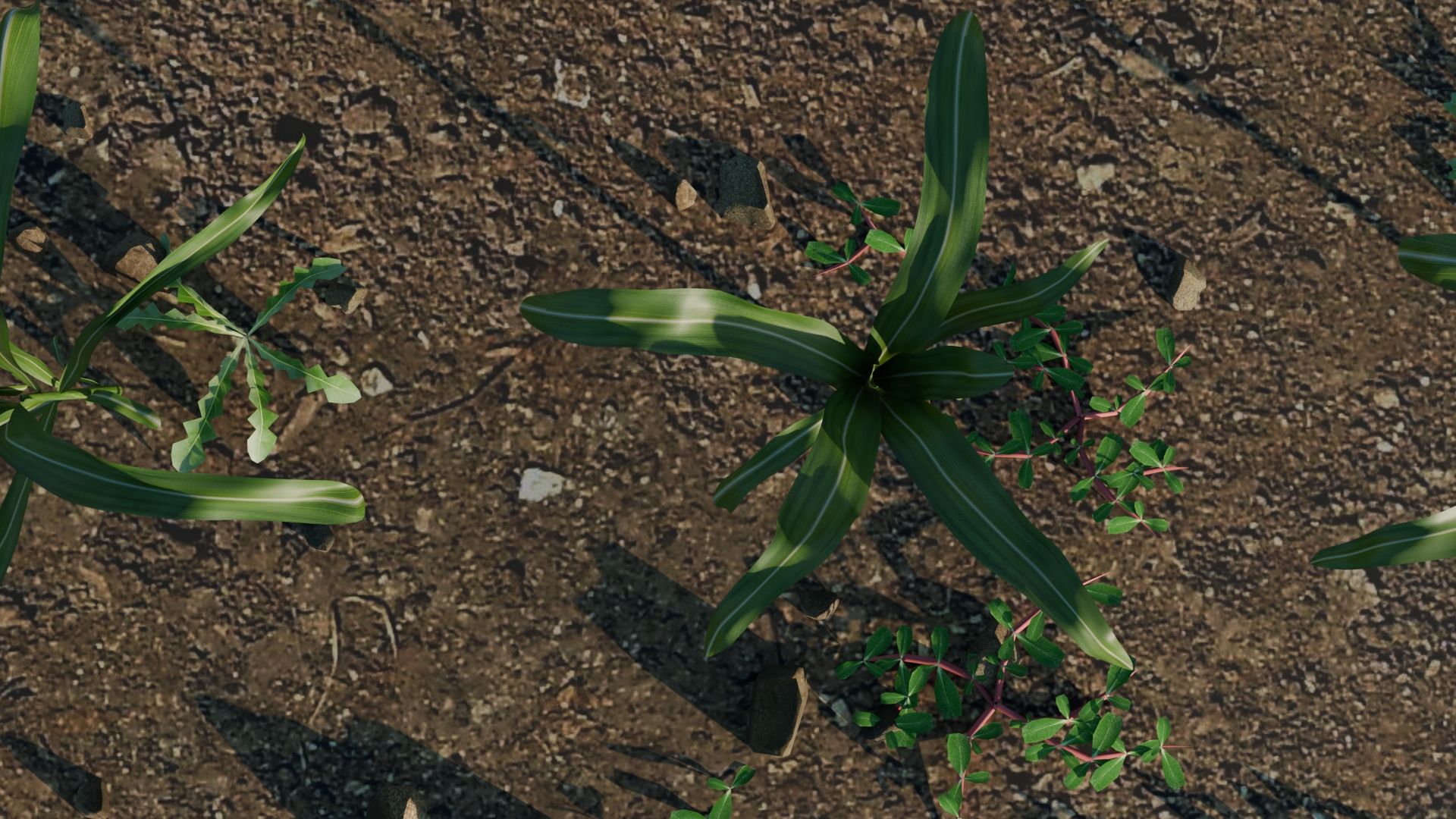}
    }%
    \hfill
    \subcaptionbox{Afternoon, small maize, medium weed density, 30° camera angle\label{fig:b}}[0.48\columnwidth]{
        \includegraphics[width=\linewidth]{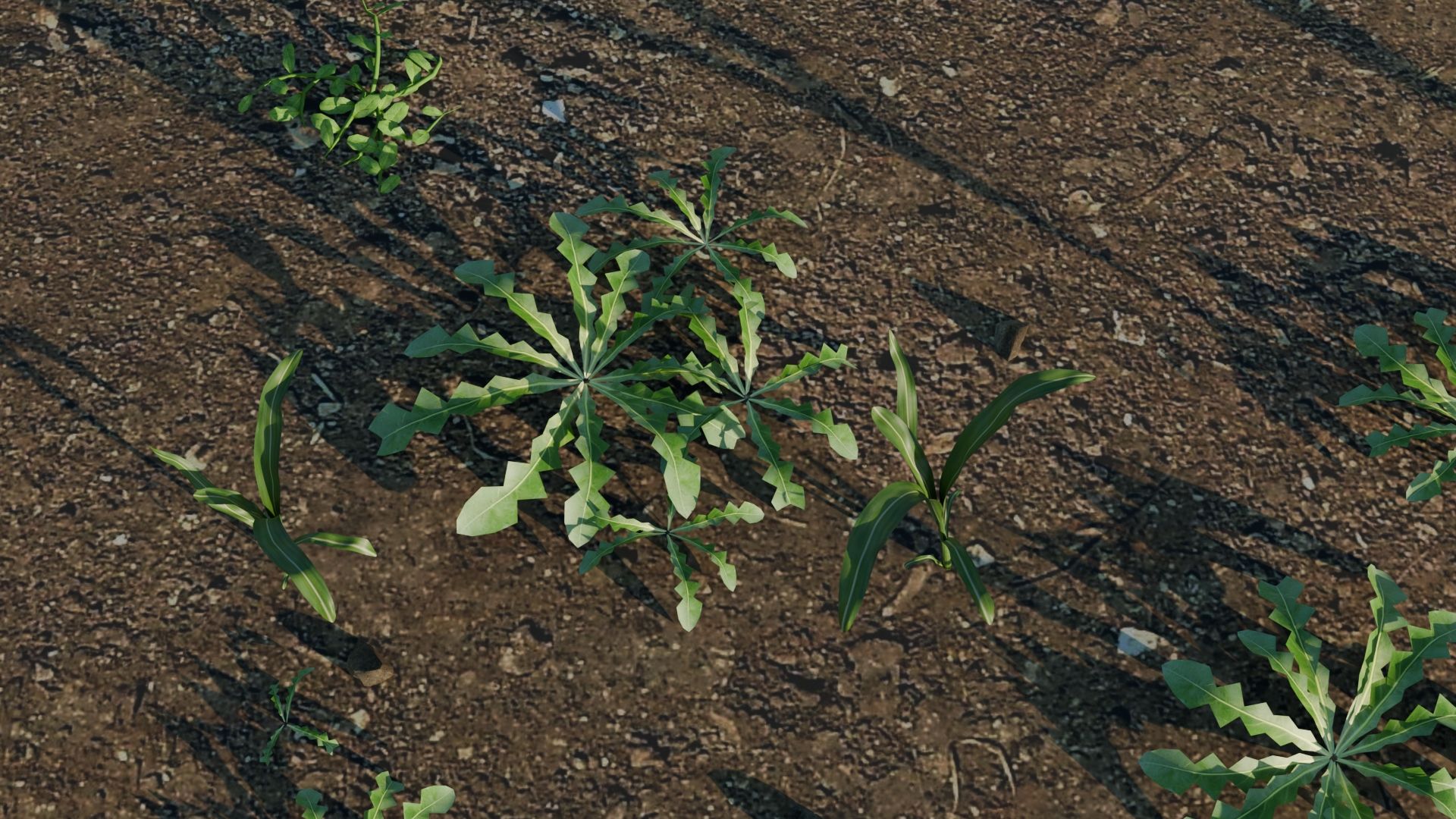}
    }

    \subcaptionbox{Noon, tall maize, high weed density\label{fig:c}}[0.48\columnwidth]{
        \includegraphics[width=\linewidth]{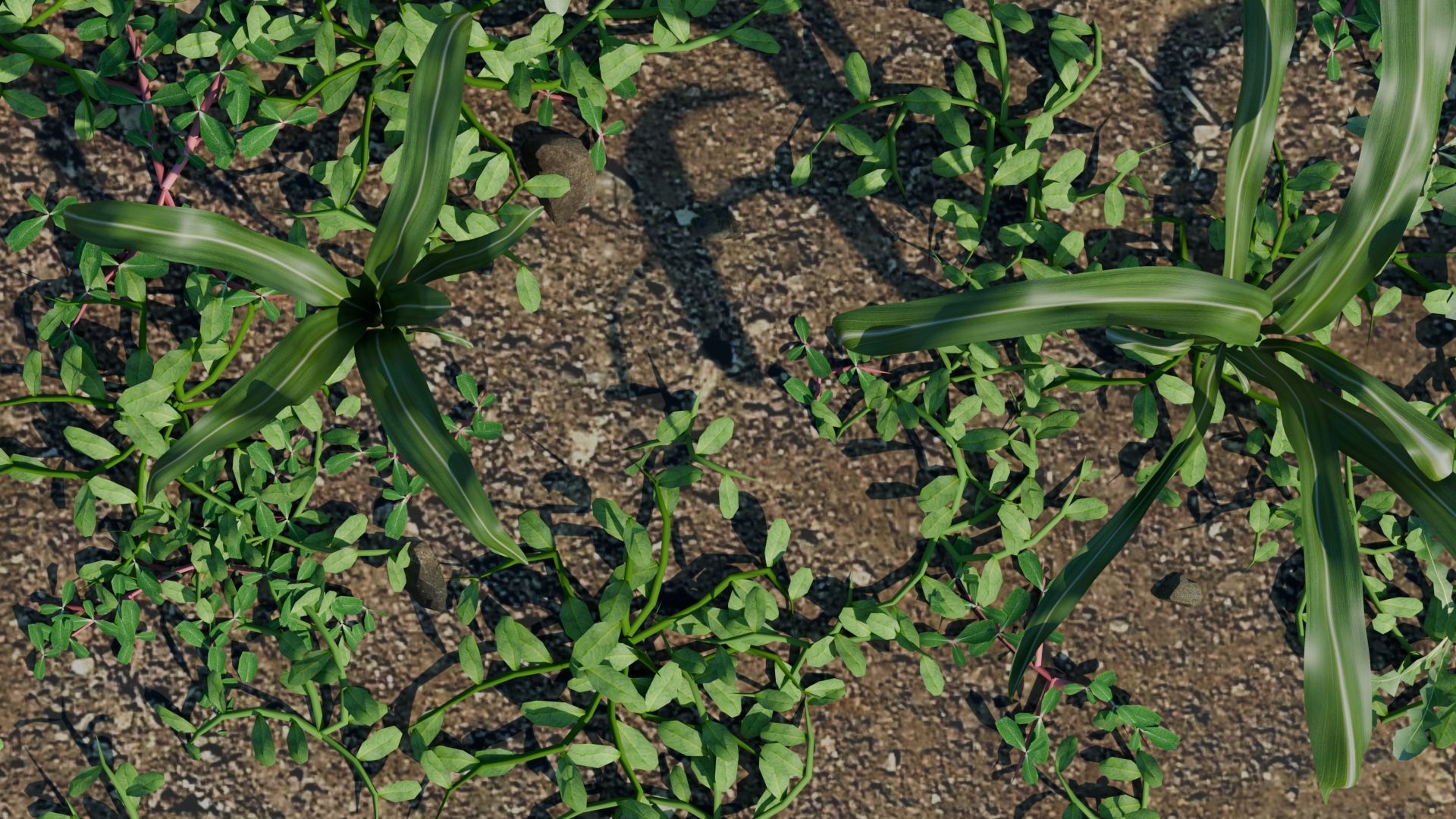}
    }%
    \hfill
    \subcaptionbox{Night, small maize, medium weed density\label{fig:d}}[0.48\columnwidth]{
        \includegraphics[width=\linewidth]{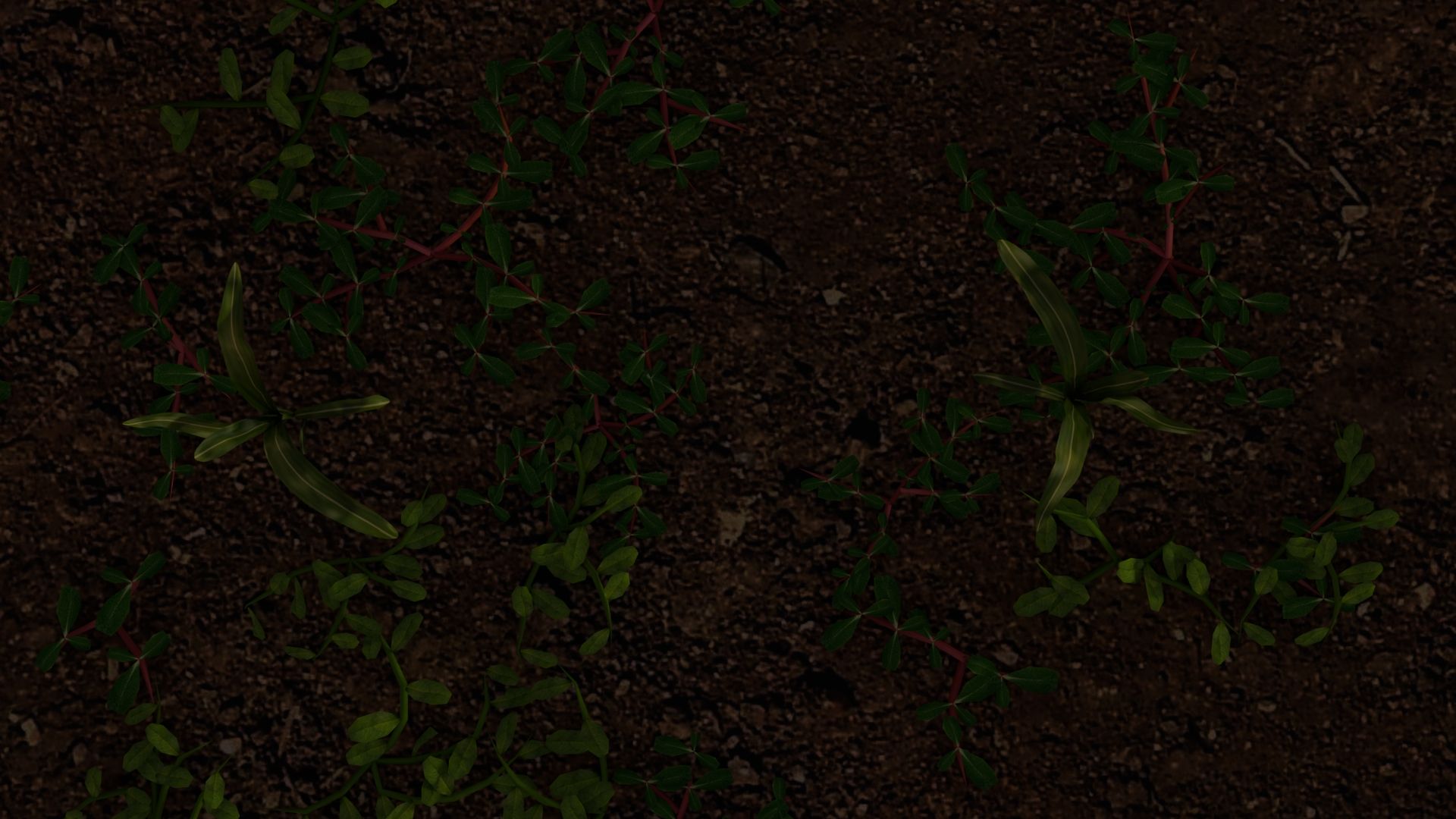}
    }

    \caption{Examples of synthetic images generated with CropCraft under different conditions of time of day, maize growth stage, weed density, and camera angle.}
    \label{fig:synthetic_examples}
\end{figure}

We generated synthetic images along with their corresponding annotated versions using CropCraft.
In this study, we focused on maize (\textit{Zea mays}) as the target crop, although other plant models are available in CropCraft.
For weeds, we used three species in the simulations: \textit{Portulaca oleracea}, \textit{Polygonum aviculare}, and \textit{Taraxacum officinale}.
As part of this work, a dataset of 10,000 synthetic images was generated using CropCraft and is made publicly available at \url{https://doi.org/10.57745/TNCSLP}.
For the experiments reported in this paper, a subset of 1500 images was used, selected to cover a representative range of conditions.
We varied maize plant height (small, medium, large), weed density (low, medium, high), time of day (noon, afternoon, night), and camera angle (top-down view, $30^{\circ}$ rotation on the roll).
This resulted in six subsets of synthetic images, each consisting of 250 images and their corresponding annotations.
The resolution of the synthetic images is 1920x1080 (width x height).
The annotations are semantic segmentation masks with three classes: crop, weeds, and background.
Figure \ref{fig:synthetic_examples} illustrates some examples of the generated synthetic images.

\begin{figure}[t]
    \centering
    \begin{minipage}{0.48\columnwidth}
        \centering
        \includegraphics[width=0.9\linewidth]{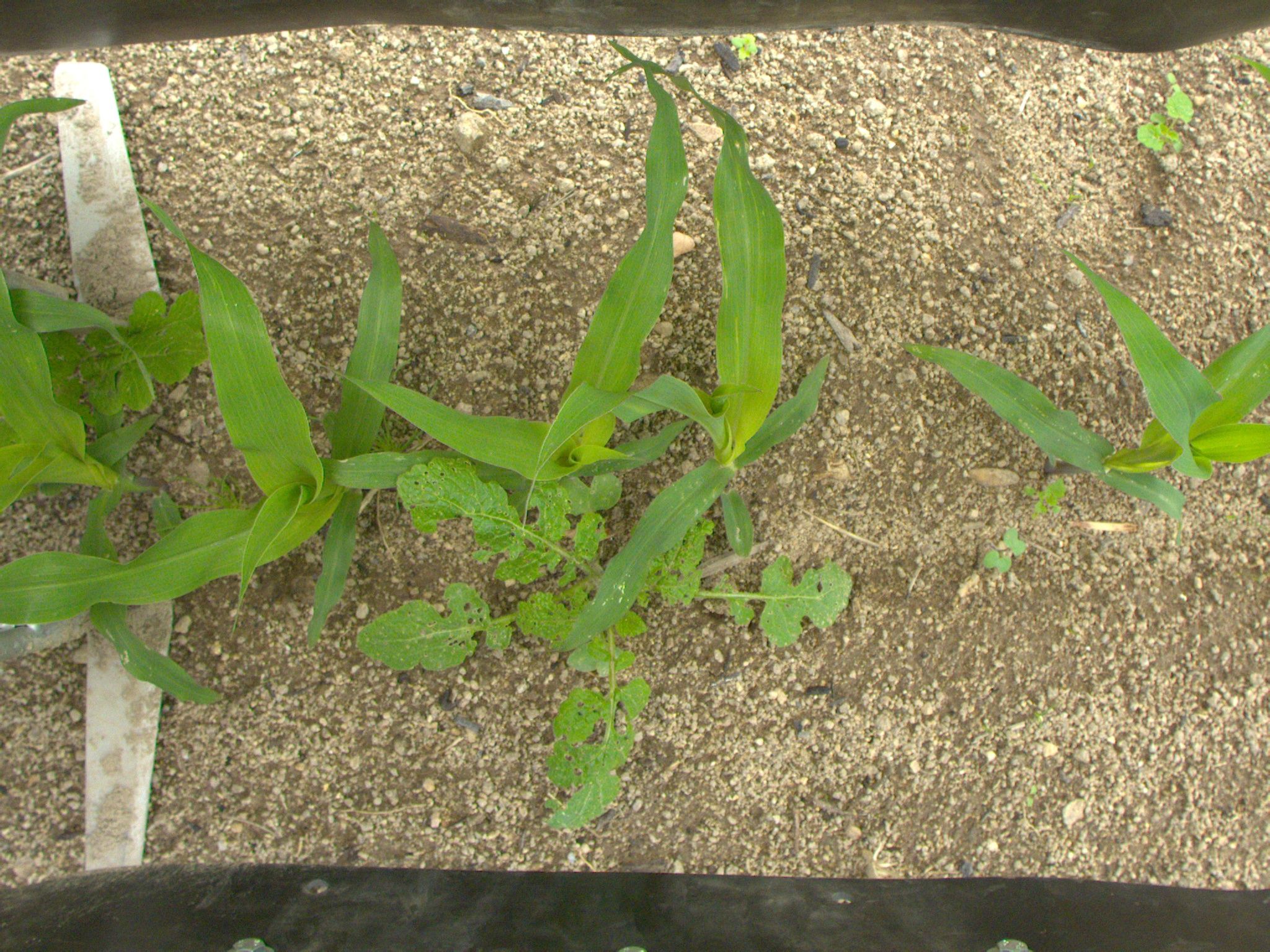}
    \end{minipage}%
    \hfill
    \begin{minipage}{0.48\columnwidth}
        \centering
        \includegraphics[width=0.9\linewidth]{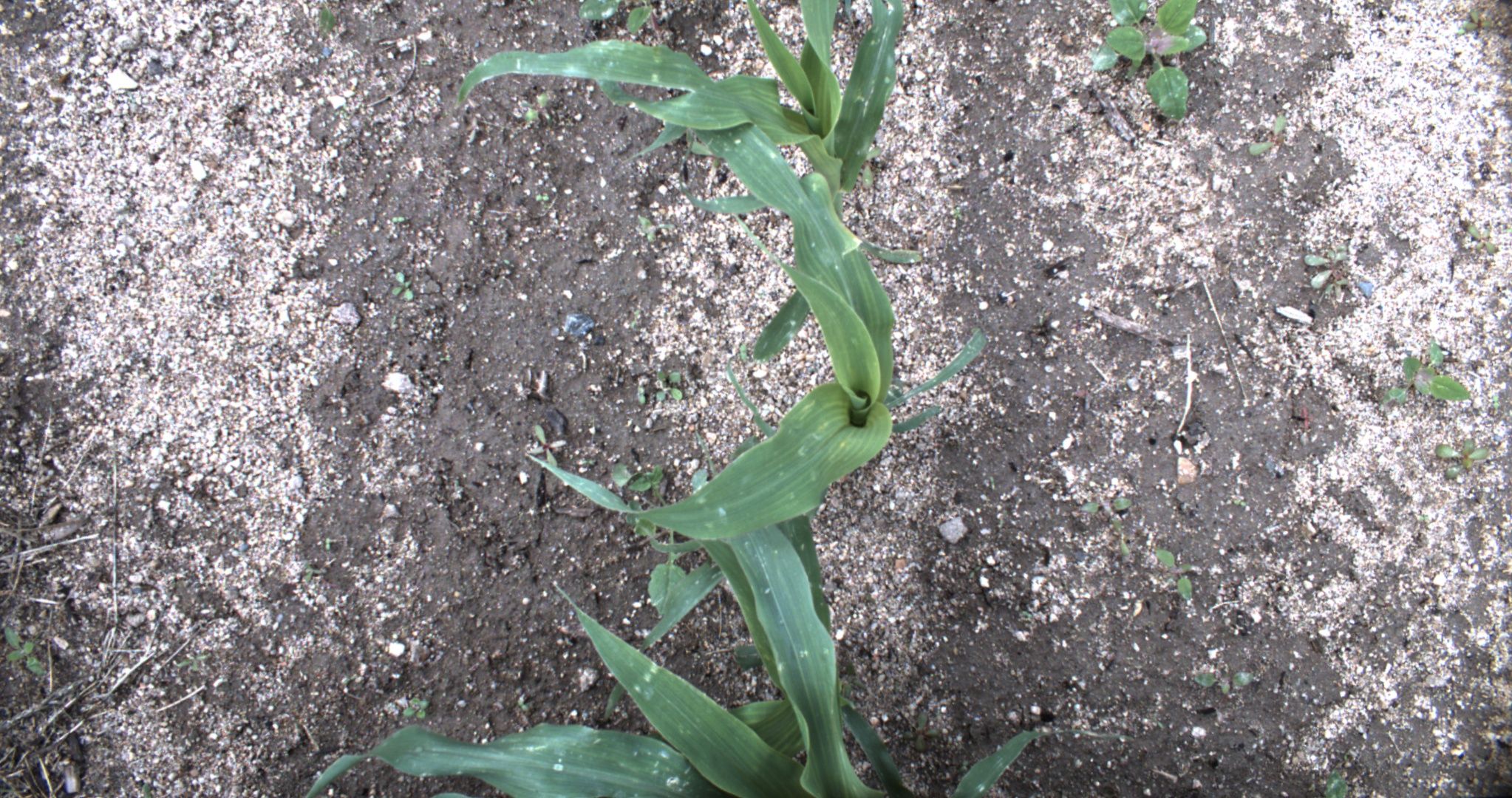}
    \end{minipage}

    \vspace{1ex} 

    \begin{minipage}{0.48\columnwidth}
        \centering
        \includegraphics[width=0.9\linewidth]{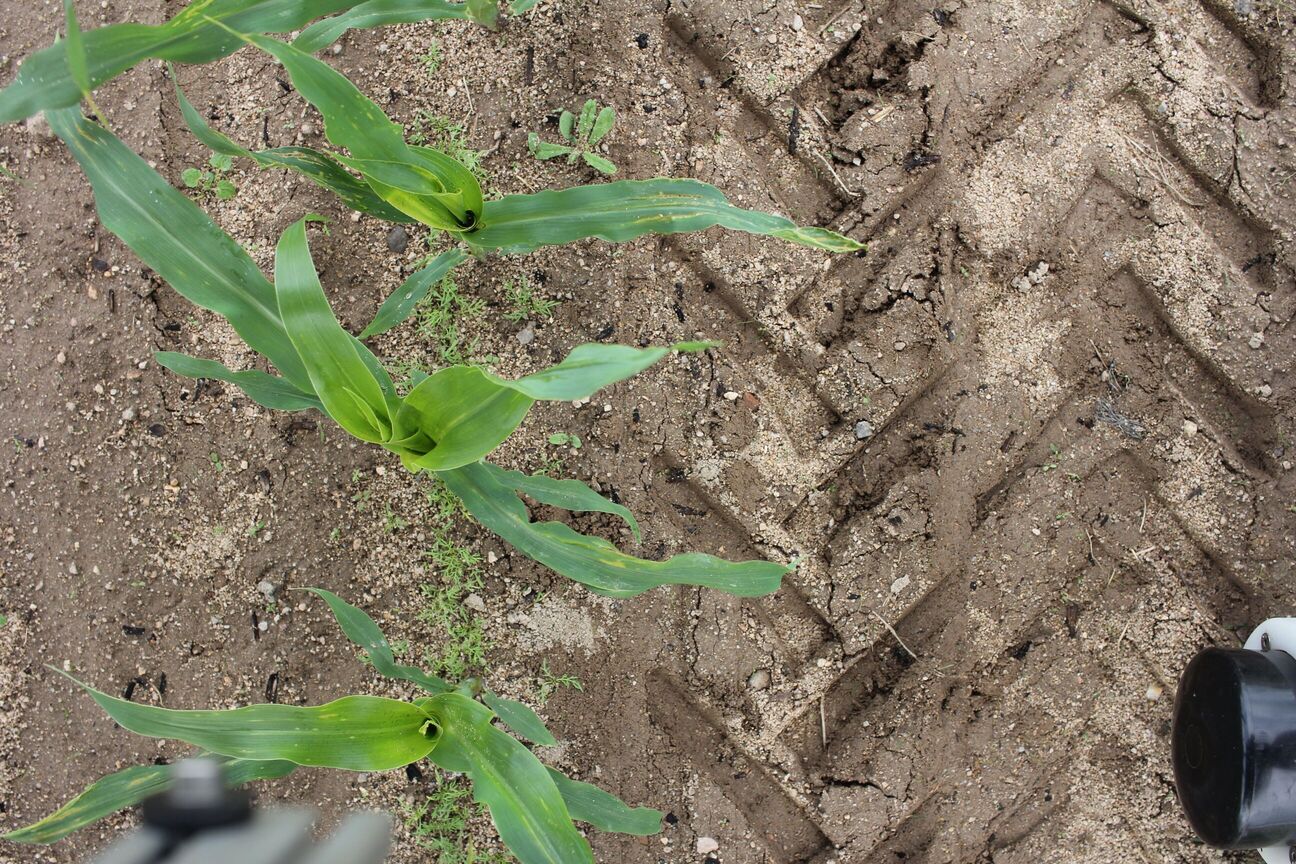}
    \end{minipage}%
    \hfill
    \begin{minipage}{0.48\columnwidth}
        \centering
        \includegraphics[width=0.9\linewidth]{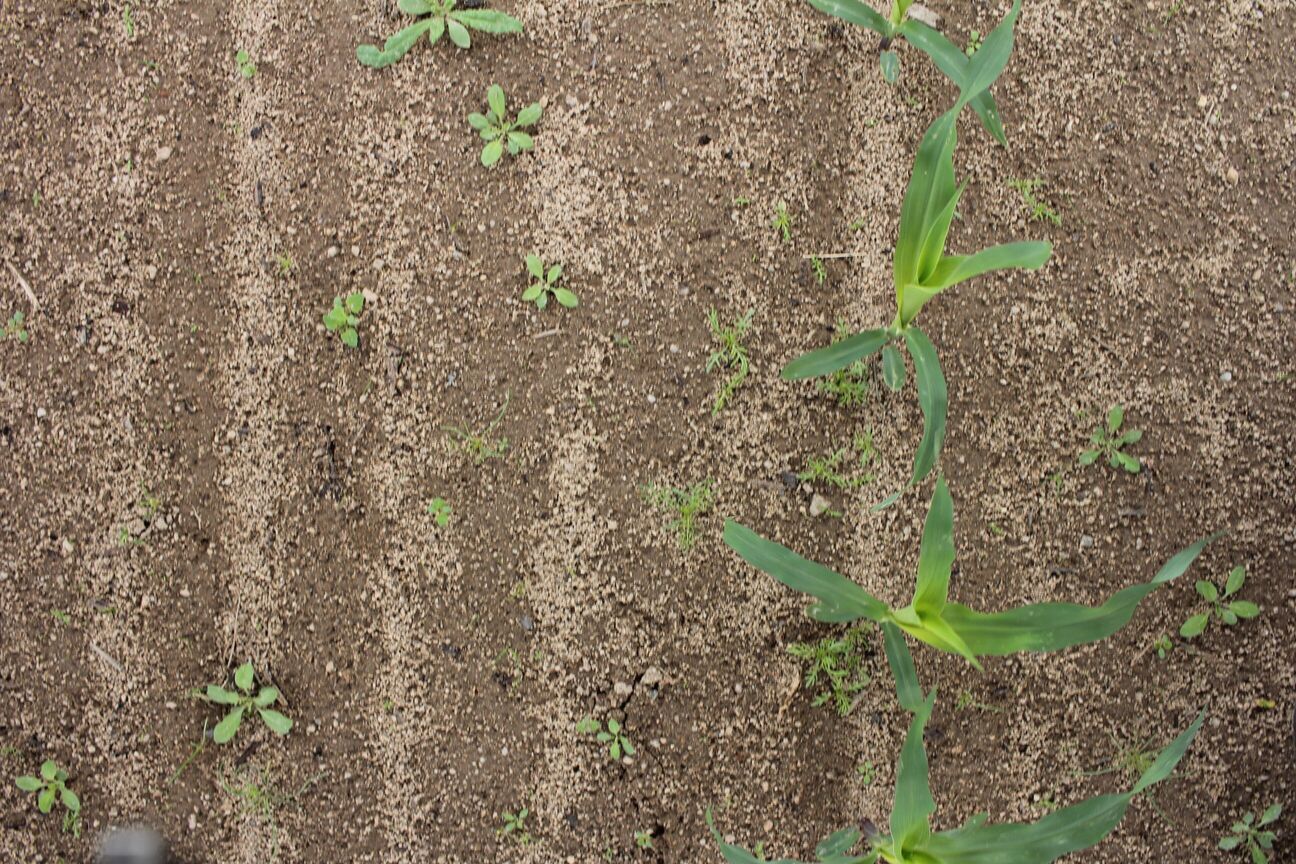}
    \end{minipage}

    \caption{Examples of real images collected in maize fields during the ROSE and ACRE challenges. The datasets were acquired in different years with different robots and cameras.}
    \label{fig:real_examples}
\end{figure}

\subsubsection{Real Data}
To evaluate the accuracy of models trained on synthetic images, we used two datasets of real images captured by robots in agricultural fields.
The first dataset comes from the ROSE \cite{avrin_design_2020} and ACRE \cite{bertoglio_design_2021, bertoglio_acre_2025} challenges, where autonomous weeding robots were benchmarked during real field campaigns.
It consists of 1000 images collected in Montoldre, France, depicting maize plants together with several weed species, including \textit{Lolium perenne}, \textit{Sinapis arvensis}, \textit{Matricaria chamomilla}, and \textit{Chenopodium album}.
An important point to note is that the weed species present in the real images are not the same as those used in the synthetic dataset, since these species are not yet available in CropCraft.
The real images were acquired by three different robots, using different cameras, and across different years (2019, 2021, and 2022).
Image resolutions vary across subsets, ranging from 1024x768 (lowest) to 5184x3456 (highest).
All images were manually annotated with semantic segmentation labels distinguishing crop, weeds, and background. This dataset will hereafter be referred to as \textit{Real Montoldre}.
Examples of these images are shown in Figure \ref{fig:real_examples}.

To further test the generalization of our models, we also included another public dataset, namely the CropAndWeed dataset \cite{steininger_cropandweed_2023}.
From this dataset, we selected only the maize images, which contain multiple weed species such as \textit{Portulaca oleracea}, \textit{Polygonum aviculare}, \textit{Sinapis arvensis}, \textit{Matricaria chamomilla}, and \textit{Chenopodium album}.
A total of 1836 maize images were used, with a resolution of 960x544.
This dataset is diverse in terms of lighting conditions, plant sizes, soil appearance, and weed densities.
It will hereafter be referred to as \textit{Real Steininger}.

The number of pixels per class in the three datasets is as follows (in millions): for the synthetic data, 2627M background, 191M crop, and 291M weed; for the \textit{Real Montoldre} dataset, 5177M background, 679M crop, and 219M weed; and for the \textit{Real Steininger} dataset, 3731M background, 75M crop, and 29M weed.

\subsection{Experimental Setup}\label{sec:exp_setup}
The task addressed in this study is semantic segmentation, i.e., classifying each pixel of an input image into one of three classes: crop, weed, or background.
To this end, we first trained several semantic segmentation models with different encoders on the synthetic dataset in order to identify the most suitable configuration for our use case.
Model performance was evaluated using the Intersection over Union (IoU), defined as the ratio between the area of overlap and the area of union between the predicted segmentation and the ground truth.
Specifically, we report the mean IoU (mIoU) over the three classes, as well as the IoU for the crop class and the weed class separately.
Once the best-performing model was identified, it was trained and tested on different combinations of datasets.
Regarding the training sets, we experimented with three strategies:
\begin{itemize}
    \item training solely on synthetic data,
    \item training solely on real data,
    \item training on a mix of real and synthetic data.
\end{itemize}
In particular, two mixed datasets were constructed by varying the number of real images included (20 and 700).
This allowed us to investigate the effect of gradually increasing the amount of real data, from very small to relatively large subsets.
A similar analysis was carried out with the \textit{Real Montoldre} dataset, where the model was trained exclusively on subsets of 20 and 700 real images to assess the added value of synthetic data.

\begin{table*}[t]
\centering
\small
\setlength{\tabcolsep}{5pt}
\caption{Test IoU performance of different segmentation models trained on synthetic data.}
\label{tab:segmentation_results}
\begin{tabularx}{\textwidth}{lcX|ccc}
\toprule
\textbf{Model} & \textbf{\#Params (M)} & \textbf{Encoder} & \multicolumn{3}{c}{\textbf{IoU (\%)}} \\
\cmidrule(l){4-6}
 & & & \textbf{Avg} & \textbf{Crop} & \textbf{Weed} \\
\midrule
DeepLabV3+  & 26.7  & ResNet-50     & 94.1 & 95.4 & 87.7 \\
SegFormer   & 87.6  & SwinV2-B      & 93.6 & 94.4 & 87.2 \\
SegFormer   & 88.3  & ConvNeXt-B    & 95.3 & \textbf{96.9} & 89.7 \\
DeepLabV3+  & 89.4  & ConvNeXt-B    & 94.9 & 96.0 & 89.5 \\
UPerNet     & 97.7  & ConvNeXt-B    & \textbf{95.8} & \textbf{96.9} & \textbf{91.2} \\
DPT         & 107.8 & SwinV2-B      & 94.1 & 95.2 & 87.8 \\
\bottomrule
\end{tabularx}
\end{table*}

\begin{table*}[ht]
\centering
\small
\setlength{\tabcolsep}{3pt}
\caption{Test IoU performance of models trained on different datasets and tested across domains. Best results for each real dataset are highlighted in bold, while the second-best results are italicized. The last column reports the average mIoU computed between the two real datasets (\textit{Real Montoldre} and \textit{Real Steininger}).}
\label{tab:cross_domain_results}
\begin{adjustbox}{max width=\textwidth}
\begin{tabularx}{\textwidth}{l|YYY|YYY|YYY|Y}
\toprule
\multirow{2}{*}{\textbf{Training Mode}} & 
\multicolumn{3}{c|}{\textbf{Synthetic}} & 
\multicolumn{3}{c|}{\textbf{Real Montoldre}} & 
\multicolumn{3}{c|}{\textbf{Real Steininger \cite{steininger_cropandweed_2023}}} &
\multirow{2}{*}{\textbf{Avg Real}} \\
\cmidrule(lr){2-4} \cmidrule(lr){5-7} \cmidrule(lr){8-10}
& \textbf{Avg} & \textbf{Crop} & \textbf{Weed} 
& \textbf{Avg} & \textbf{Crop} & \textbf{Weed} 
& \textbf{Avg} & \textbf{Crop} & \textbf{Weed} \\
\midrule
Only synthetic data (1050 images)   & 96.1 & 97.5 & 91.9 & 70.0 & 76.1 & 38.5 & 73.9 & 73.1 & 49.4 & 72.0 \\
Synthetic + 20 Real M. images       & -- & -- & -- & 76.3 & 83.7 & 50.4 & 73.2 & 70.6 & 49.7 & 74.8 \\
Synthetic + 700 Real M. images      & -- & -- & -- & \textit{80.5} & \textit{85.6} & \textit{59.9} & \textit{76.9} & \textit{76.3} & \textit{55.1} & \textbf{78.7} \\
Real Montoldre 20 images            & -- & -- & -- & 77.8 & 84.9 & 52.4 & 63.8 & 64.6 & 27.6 & 70.8 \\
Real Montoldre 700 images           & -- & -- & -- & \textbf{80.6} & \textbf{86.1} & \textbf{59.8} & 73.8 & 73.3 & 48.9 & \textit{77.2} \\
Real Steininger 1285 images       & -- & -- & -- & 57.2 & 64.9 & 18.4 & \textbf{83.1} & \textbf{87.8} & \textbf{62.2} & 70.2 \\
\bottomrule
\end{tabularx}
\end{adjustbox}
\end{table*}

\subsection{Deep Learning Models Training}\label{sec:dl_training}
Each dataset was randomly split into training, validation, and test sets, with proportions of 70\%, 10\%, and 20\%, respectively.
For the \textit{Real Montoldre} dataset (1000 images), the training portion (700 images) was further split to create a subset of 20 images used in the experiments described above.
All models were trained under the same hyperparameter settings on a single NVIDIA Tesla V100 GPU with 32 GB of VRAM.
Training was performed for 5k iterations using the AdamW optimizer, with a learning rate of 0.001 and a weight decay of 0.001.
A cosine annealing learning rate scheduler and a weighted cross-entropy loss were employed, with class weights of 1, 2, and 4 for background, crop, and weed, respectively.
The batch size was set to 8.
During training, images were augmented using a combination of transformations, including random resizing, random cropping, horizontal and vertical flips, shifting/scaling/rotating, color jittering, random brightness and contrast adjustments, Gaussian blur, and Gaussian noise.
As mentioned, we tested multiple architectures and encoders to identify the best-performing setup.
The models evaluated were DeepLabv3+, SegFormer, UPerNet, and DPT, with encoders ResNet-50, SwinV2-B, and ConvNeXt-B.
During training, images were randomly cropped to 384x384 pixels.
For testing, images were resized and padded to 1536x1536 pixels to enable a fair comparison across models, given the size constraints of the SwinV2 encoder.
In the remaining experiments, we used an input size of 1088x1920 pixels, which better matches the original aspect ratio of the images.
All results were obtained using the checkpoint from the final training iteration.
The code used for training and evaluating the models is made publicly available at \url{https://github.com/Romea/crop-weed-segmentation}.

\section{Results and Discussion}
We first evaluated several state-of-the-art semantic segmentation models on the synthetic dataset to identify the most suitable architecture.
This choice reflects a scenario where real data are scarce or unavailable.
The mIoU scores on the synthetic test set are summarized in Table \ref{tab:segmentation_results}.
Overall, the models performed similarly, with UPerNet paired with a ConvNeXt-B encoder achieving the highest accuracy.
Notably, DeepLabV3+ with a ResNet-50 encoder (27M parameters) delivered results comparable with much larger models using ConvNeXt-B or SwinV2-B encoders (90-100M parameters).
This suggests that ResNet-50 offers a practical trade-off between performance and efficiency.
Since UPerNet produced the best results overall, we selected it for the subsequent experiments.

We then examined how models trained on synthetic images perform when tested on real datasets.
The results, shown in Table \ref{tab:cross_domain_results}, confirm a clear performance drop when models trained exclusively on synthetic data are applied to real images.
This degradation is mainly due to differences in the visual appearance of plants, particularly in shape and texture, between synthetic and real domains.
Specifically, mIoU decreased by 10.6\% on the \textit{Real Montoldre} dataset and by 9.2\% on \textit{Real Steininger}, thus leading to an average sim-to-real performance gap of 9.9\%.
The decline was especially pronounced for the weed class, likely because the weed species represented in the synthetic and real datasets are not the same.
Adding real images to the synthetic training set improved performance on \textit{Real Montoldre}, with accuracy increasing as more real images were included.
Remarkably, even with only 20 real images, weed IoU improved by 11.9\%.
This suggests that small amounts of real data complement the knowledge learned from synthetic images.
However, when training solely on these real subsets (without synthetic data), results were superior.
This indicates that to maximize performance on a specific dataset/domain, synthetic data provide little benefit compared to real data.

On the other hand, synthetic images appear valuable for improving generalization.
For instance, training only on a small subset of 20 \textit{Real Montoldre} images resulted in weaker performance on \textit{Real Steininger} compared to training on synthetic data.
With a larger subset of 700 Montoldre images, cross-domain performance was comparable between real and synthetic training (73.8 for real data and 73.9 for synthetic), indicating that sufficiently large real datasets can also support generalization.
However, models trained on \textit{Real Steininger} generalized poorly to \textit{Real Montoldre}, despite \textit{Real Steininger} being the larger dataset.
In this case as well, training only on synthetic data led to better performance.
Moreover, the second-best results on both real datasets were obtained by training on a mix of synthetic and real data (Synthetic + 700 Real M. images), suggesting that synthetic images can enhance the robustness of models across domains, even when real data are available.

\section{Conclusion}
	This paper presented CropCraft, an open-source procedural world generator designed to support the development and evaluation of robotic systems for agricultural applications.
CropCraft produces 3D simulation environments from a simple YAML configuration file, enabling users to model a wide range of agroecological scenarios, including intercropping, vineyards, and weed-infested fields, without requiring expertise in 3D modeling software.
The framework relies on stochastic placement algorithms to introduce the spatial variability inherent to real agricultural fields: Gaussian noise on plant poses, log-normal variation in plant size, configurable probabilities of missing plants, and density-map-based weed scattering using Poisson disk sampling modulated by heterogeneous terrain noise.
In addition to the 3D geometry, CropCraft produces complementary ground-truth data in the form of structured field description files, lidar intensity annotations for Gazebo simulation, and pixel-accurate semantic segmentation masks for camera-based datasets.

The utility of the framework was demonstrated through an application to crop-weed semantic segmentation using deep learning.
A dataset of 10,000 synthetic images of maize fields was generated with CropCraft, covering a range of weed densities, plant growth stages, and lighting conditions.
Several state-of-the-art segmentation architectures were trained and evaluated in a sim-to-real transfer setting.
The best-performing model, UPerNet with a ConvNeXt-B encoder, achieved a sim-to-real gap of approximately 10\% mIoU, outperforming previous synthetic generation approaches that reported gaps of around 20\%.
Furthermore, the experiments showed that incorporating even a few real images alongside synthetic data consistently improves generalization across domains, offering practical guidance for training strategies when real annotated data are scarce.

Multiple directions remain open for future work.
First, the current plant library covers a limited set of species; expanding it with additional crop types, weed species, and growth stages would increase the breadth of scenarios that can be simulated and would help reduce the domain gap caused by species mismatch between synthetic and real datasets.
Second, CropCraft does not currently model vegetation deformation, such as crops bent by robot passage or plants flattened by wind.
Incorporating physics-based or procedural deformation models would improve the realism of dynamic simulation scenarios and benefit the development of contact-aware navigation algorithms.
Third, while the Gazebo simulator is the primary export target, extending support to other platforms such as Isaac Sim would broaden the range of usable physics engines and rendering pipelines.
Fourth, the ground-truth production pipeline could be extended to support instance segmentation masks and 3D bounding box annotations, which are required by a growing number of perception algorithms.
Finally, the integration of soil and terrain variability, including surface roughness, furrows, and moisture-dependent appearance, would further increase the fidelity of the generated environments, particularly for the development of robust navigation and soil-interaction algorithms.

\section*{CRediT authorship contribution statement}
Riccardo Bertoglio: Conceptualization, Methodology, Software, Validation, Investigation, Data Curation, Writing - Original Draft, Writing - Review \& Editing, Visualization.
Cyrille Pierre: Conceptualization, Software, Writing - Original Draft, Visualization.
Johann Laconte: Conceptualization, Software, Writing - Original Draft, Writing - Review \& Editing.
Roland Lenain: Conceptualization, Project administration, Funding acquisition, Writing - Review \& Editing, Visualization.

\section*{Declaration of competing interest}
The authors declare that they have no known competing financial interests or personal relationships that could have appeared to influence the work reported in this paper.

\section*{Acknowledgments}
This work was financed in the framework of NINSAR project by ANR (PEPR Agroécologie et Numérique) under France 2030 (grant ANR-22-PEAE-0007).
Moreover, this work has been partially supported by ROBOTEX 2.0, the French Infrastructure in Robotics under the grants ROBOTEX (EQUIPEX ANR-10-EQPX-44-01) and TIRREX (EQUIPEX+ grant ANR-21-ESRE-0015).
This work has also received the support of the French government research program "Investissements d'Avenir" through the IDEX-ISITE initiative 16-IDEX-0001 (CAP 20-25), the IMobS3 Laboratory of Excellence (ANR-10-LABX-16-01).
We are grateful to the Mésocentre Clermont-Auvergne of the Université Clermont Auvergne for providing access to computing and storage resource facilities.

\section*{Data availability}
An extended synthetic dataset of 10,000 images of maize fields with varying weed densities, growth stages, and lighting conditions generated for the application section is available at \url{https://doi.org/10.57745/TNCSLP}.
The ROSE dataset composing the set of Real Montoldre images used for evaluation is available upon request.
The ACRE Crop-Weed Dataset composing the Real Montoldre images used for evaluation is available at \url{https://doi.org/10.5281/zenodo.8102217}.
The CropAndWeed Dataset from which we extracted the images composing the set of Real Steininger images is available at \url{https://github.com/cropandweed/cropandweed-dataset}.

\bibliographystyle{elsarticle-num-names}
\bibliography{CropCraft}

\end{document}